\documentclass{article}

\usepackage{microtype}
\usepackage{graphicx}
\usepackage{subcaption}
\usepackage{booktabs} 

\usepackage[usenames,dvipsnames]{xcolor}
\usepackage{hyperref}

\newcommand{\method}{TACO}

\usepackage[accepted]{icml2018}

\icmltitlerunning{Temporal Alignment for Control}

\usepackage{mathtools}
\usepackage{amsmath}
\usepackage{color}
\usepackage{amssymb}
 
\DeclareMathOperator*{\argmax}{\arg\!\max}

\iftrue
\iftrue 
    
\fi

\fi

\begin{document}

\twocolumn[
\icmltitle{TACO: Learning Task Decomposition via Temporal Alignment for Control}





\begin{icmlauthorlist}
\icmlauthor{Kyriacos Shiarlis}{ams}
\icmlauthor{Markus Wulfmeier}{ox}
\icmlauthor{Sasha Salter}{ox}
\icmlauthor{Shimon Whiteson}{oxc}
\icmlauthor{Ingmar Posner}{ox}
\end{icmlauthorlist}

\icmlaffiliation{ams}{Informatics Institute, University of Amsterdam, Netherlands}
\icmlaffiliation{ox}{Department of Engineering Science, University of Oxford, United Kingdom}
\icmlaffiliation{oxc}{Department of Computer Science, University of Oxford, United Kingdom}
\icmlcorrespondingauthor{Kyriacos Shiarlis}{k.c.shiarlis@uva.nl}

\icmlkeywords{Machine Learning, ICML, Learning from Demonstration, Modular Learning, Sequence Alignment }

\vskip 0.3in
]



\printAffiliationsAndNotice{} 



\begin{abstract}

Many advanced \emph{Learning from Demonstration} (LfD) methods consider the decomposition of complex, real-world tasks into simpler sub-tasks. By reusing the corresponding sub-policies within and between tasks, they provide training data for each policy from different high-level tasks and compose them to perform novel ones. However, most existing approaches to modular LfD focus either on learning a single high-level task or depend on domain knowledge and temporal segmentation. By contrast, we propose a weakly supervised, domain-agnostic approach based on \emph{task sketches}, which include only the sequence of sub-tasks performed in each demonstration. Our approach simultaneously aligns the sketches with the observed demonstrations and learns the required sub-policies, which improves generalisation in comparison to separate optimisation procedures.
We evaluate the approach on multiple domains, including a simulated 3D robot arm control task using purely image-based observations. The approach performs commensurately with fully supervised approaches, while requiring significantly less annotation effort, and significantly outperforms methods which separate segmentation and imitation.

\vspace{-0.3cm}


\end{abstract}

%

\section{Introduction}

\emph{Learning from demonstration}~(LfD) represents a popular paradigm to teaching complex behaviours to robots and virtual agents through demonstrations, without the need for explicit programming or other description of a task, such as a cost function \citep{argall2009survey}. 
The benefits of LfD over manual task definitions are numerous. First, some behaviours are difficult to program or manually encode, but can be easily demonstrated. Second, while programming behaviours requires expert knowledge of the application platform, LfD requires only the ability to control the agent. 

However, complex real world tasks pose a severe challenge for simple LfD approaches such as \emph{behavioural cloning} (BC). While a complex task can often be broken down into simpler sub-tasks, the algorithm itself lacks the means to discover the decomposition and must instead learn a monolithic policy for the whole task. The resulting policies are not only more complex but also less reusable. For example, a cube stacking task can be broken down into sub-tasks for approaching a cube, grasping it, moving to a location, and placing it onto an existing stack. In particular, sub-policies for approaching and grasping the cube can be reused in other manipulation routines such as rotating or throwing the cube.


\emph{Modular LfD} addresses this challenge by modelling the task as a composition of sub-tasks for which reusable sub-policies (modules) are learned.  These sub-policies are often easier to learn and can be composed in different ways to execute new tasks, enabling zero-shot imitation.

One approach to modular LfD is to provide the learner with additional information about the demonstrations. This comes in many forms, e.g., manual segmentation of demonstrations \citep{hovland1996skill}, interactive feedback during learning \citep{niekum2015learning}, or prior knowledge about the task. Such domain knowledge may come in the form of motion primitives, hard-coded parametric motion models \citep{ekvall2006learning,schaal2006dynamic} and problem-specific state modelling \citep{abdo2013learning,cst2012}. An additional benefit is that since these methods ground demonstrations on human-defined sub-tasks, the learned policies are interpretable.
They are, however, labour intensive, perform sub-optimally if the underlying assumptions are incorrect, and require domain knowledge. 


In this work, we consider a general, weakly supervised, modular LfD setting where demonstrations are augmented only with a \emph{task sketch}. This sketch describes the sequence of \emph{sub-tasks} that occur within the demonstration, without additional information on their alignment 
(see Figure \ref{fig:mlfd}). 


Drawing inspiration from speech recognition \citep{graves2006connectionist} as well as modular reinforcement learning \citep{andreas2017modular}, we introduce \emph{temporal alignment for control} (\method)\footnote{The \method~algorithm and experiments will soon be available at \href{https://sites.google.com/view/taco-ml}{https://sites.google.com/view/taco-ml}.}, an efficient, domain agnostic algorithm that learns modular and grounded policies from high level task descriptions, while relying purely on weak supervision. 

Instead of considering the alignment of a demonstration with the task sketch and the learning of associated sub-policies as two separate processes, in \method~the imitation learning stage affects the alignment and vice-versa by maximising the joint likelihood of the observed sketch and the observed action sequence given the states.
\method~learns one sub-policy for each sub-task present in the data and extends each sub-policy's action space to enable self-termination. 
At test time, the agent is presented with new, potentially unseen, and often longer sketches, which it executes by composing the required sub-policies. 
In addition to simplifying the imitation of complex tasks, the approach enables zero-shot imitation given only a sketch. 

We evaluate the performance of \method~on four domains of varying complexity. First, we consider two toy domains, a 2D navigation task and the Craft domain proposed by \citet{andreas2017modular}. 
Finally, we consider the scenario of controlling a simulated robot arm to use a number pad and extend the task to use only image-based observations.  We demonstrate that, in all domains, policies trained using \method~are capable of matching the performance of policies trained using a fully supervised method, where the segmentation of the demonstration is provided, at a small fraction of the labelling cost. At the same time, the approach significantly outperforms our baselines which separate the optimisation processes for segmentation and imitation.

\vspace{-5pt}


\section{Related Work}\label{sec:related}

Learning from demonstration encompasses a wide range of techniques that focus on learning to solve tasks based on (human) expert demonstrations  \citep{argall2009survey}. 
The fields of modular and hierarchical LfD aim to extract reusable policy primitives from complex demonstrations to increase data efficiency and transfer knowledge between tasks.

In robotics, sub-policies can be modelled as \emph{motion primitives} \citep{schaal2006dynamic} which build the foundation for various works on modular LfD e.g. \cite{motionprim_2012, manschitz2014learning,5152385}. In this context, most similar to the ideas underlying our approach is recent work based on \emph{skill trees} \citep{cst2012} and semantically grounded finite representations \citep{niekum2015learning}. However, these approaches consider separate segmentation of the trajectories and fitting of primitives and imposes stronger constraints on the type of the controllers. In contrast, our work addresses segmenting the demonstrations and learning the policies in one combined process. 

Interleaving the two processes has been shown to provide better segmentations and policies in recent work \cite{lioutikov2017learning}. The approach however considers learning via policy embeddings in task space building on probabilistic motion primitives \cite{paraschos2013probabilistic}, which restricts the approach with respect to the types of tasks and observations. The method presented in this paper is less constrained and can handle arbitrary differentiable function approximator for the control policies. 





Recent work on hierarchical LfD transfers concepts from the \emph{options} framework \citep{sutton1999between}, which models low-level policies as actions for a meta controller, to LfD \citep{fox2017multi,krishnan2017ddco,2017arXiv170906683H}. 
Generally, options serve as tools for dividing a complex task into multiple sub-policies specialised for regions in the state space. \method~differs from option discovery frameworks \cite{fox2017multi,krishnan2017ddco} during both training and inference by replacing the functionality of the meta-controller with weak supervision in the form of a sequence of symbols.
Weak supervision constrains the learned policies to follow the description in the task sketch. This prevents degenerate cases including high-frequency switching between policies common with options \citep{krishnan2017ddco} as well as the potential collapse of the meta-controller to apply only a single option.
Furthermore, by applying task sketches at test time, we enable the composition of sub-policies in unseen and longer sequences for zero-shot imitation.

Recent work on \emph{modular reinforcement learning} (RL) \citep{andreas2017modular}, introduces the notion of sketches as additional information representing the decomposition of tasks.  Similar to our work, \citet{andreas2017modular} assume that complex tasks can be broken down into sub-tasks. Our approach exploits a similar modular structure but utilises an imitation based objective that addresses the problem of aligning sequences of different lengths.

A common approach to sequence alignment in speech recognition is \emph{connectionist temporal classification} (CTC) \citep{graves2006connectionist}.  Previous extensions of CTC have been proposed to increase its flexibility by reducing the assumptions underlying the framework \citep{graves2012sequence} and exploiting structure in the input space \citep{huang2016connectionist}. In this paper, we extend CTC by combining sequence alignment and behavioural cloning. 

\vspace{-6pt}


\section{Preliminaries}\label{sec:prelims}
In this section, we introduce the required concepts and methods for the derivation of \method. 

\subsection{Behavioural Cloning}
\emph{Behavioural cloning} (BC) models LfD as a supervised learning problem, by optimising a policy $\pi$ to maximise the likelihood of the training dataset $\mathcal{D} = \{\rho_1,\rho_2,...,\rho_M\}$, where $\rho=((s_1,a_1),(s_2,a_2),...,(s_{T},a_{T}))$ is a state-action demonstration trajectory of $T$ pairs of states $s \in \mathcal{S}$ and actions $a \in \mathcal{A}$.  Let $\pi_\theta(a|s)$ be the probability of taking action $a$ in state $s$ as modelled by a policy $\pi_{\theta}$ parameterised by $\theta$.  BC performs the following optimisation:
\begin{eqnarray}
	\theta^* = \argmax_{\theta} \mathbb{E}_{\rho}[\sum_{t=1}^{T}\log  \pi_\theta(a_t|s_t)]. \label{eq:bc}
\end{eqnarray}
One drawback of BC is its susceptibility to \emph{covariate shift}, which occurs when small errors during testing cause the agent to drift away from states it encountered during training, yielding poor performance. One way to overcome this problem is using \emph{disturbances for augmenting robot trajectories} (DART) \citep{laskey2017dart}, which introduces noise in the data collection process, allowing the agent to learn actions that can recover from errors. In this paper, unless stated otherwise, we use a training approach based on DART.

The standard formulations of BC and DART, which learn only one policy per task, lack two important properties. The first is modularity: the demonstrated behaviour can have a hierarchical structure that decomposes into modules, or sub-policies.  The second is reuse: the modules can be composed in various ways to perform different tasks.
\vspace{-4pt}
\subsection{Modular Learning from Demonstration}\label{sec:mlfd2}
\begin{figure} 
    \centering
    \includegraphics[scale = 0.52]{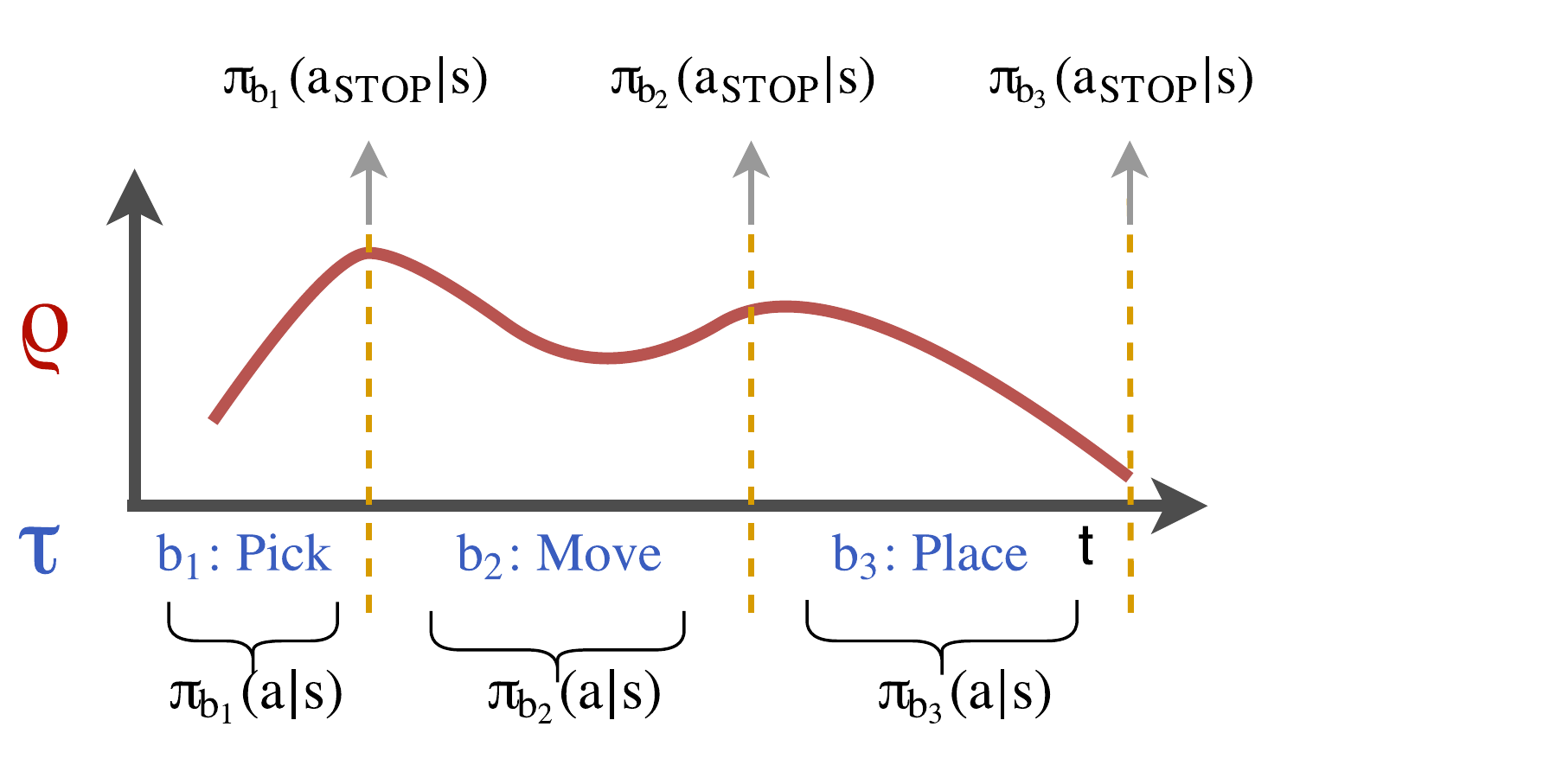}
    \vspace{ -8 mm}
    \caption{Problem setting: The trajectory $\rho$ (red) is augmented by a task sketch $\tau$ (blue). The two sequences operate at different timescales. The whole trajectory is aligned (manually or automatically) and segmented into three parts. From this alignment three separate policies are learned. The unobserved $a_{STOP}$ action for each policy is inferred to occur at the point where the policies switch from one to the other.}
    \vspace{-6pt}
    \label{fig:mlfd}
\end{figure}

Modular LfD introduces modularity and reuse to the LfD problem. A schematic is shown in Figure \ref{fig:mlfd}. To render the policies reusable, it assumes that any task can be solved by multiple sub-policies, each of which operates in an augmented action space $\mathcal{A}^+$  that includes a STOP action (i.e.,  $\mathcal{A}^+ \coloneqq \mathcal{A} \cup a_{STOP}$) that does not have to be observed in the demonstration. 
It also assumes that more than one sub-policy may be present within a demonstration. In our formulation, extra information is provided in the form of a \emph{task sketch} $\tau = (b_1,b_2,\ldots, b_{L})$, with $L \leq T$, and $b_l \in \mathcal{B}$, where $\mathcal{B} = \{1,2,\ldots,K\}$, is a dictionary of sub-tasks. The sketch indicates which sub-tasks are active in a trajectory.

Although the $a_{STOP}$ action is never observed, it can be inferred from the data. If the demonstration contains a simple task then $a_{STOP}$ is called only at the end of the demonstration. If $L = T$, then we know which policy from $\mathcal{B}$ is active at each time-step. i.e., $a_{STOP}$  for each policy is called as soon as the active policy changes within the demonstration. If all extra information is available, we can perform behavioural cloning, with two differences. First,  maximising the likelihood takes place assuming an action-augmented policy $\pi(a^+|s)$. Second, we learn $|\mathcal{B}| = K$ modular policies $\pi_{\theta_k}$ from $K$ datasets $\mathcal{D}_{k}$ containing trajectories $\rho_k$ as segmented based on $\tau$: 
\begin{eqnarray}
	\theta_{k=1,...,K}^* = \argmax_{\theta_{k}} \mathbb{E}_{\rho_k}[\sum_{t=1}^{T_{\rho}}\log  \pi_{\theta_k}(a^+_t|s_t)] \label{eq:mLfD2}. 
\end{eqnarray}
%
e

However, the fully supervised approach is labour intensive as each trajectory must be manually segmented into sub-policies. In this paper, we consider cases where $L<<T$ and $\tau$ contains only the sequence of active sub-tasks in the order they occur, without duplicates. Inferring when $a_{STOP}$ occurs is therefore more challenging as $\tau$ and $\rho$ operate at different timescales and must first be aligned.

\subsection{Sequence Alignment}\label{sec:sequence}

Since $L<<T$, we cannot independently maximise the likelihood of active sub-tasks in $\tau$ for every time step in $\rho$. The problem of aligning sequences of different lengths, a common challenge in speech and action recognition, is often addressed via \emph{connectionist temporal classification} (CTC) \citep{graves2006connectionist}. Here, we review a variant of CTC adapted to the notation introduced so far, which does not include gaps between the predictions, based on the assumption that sub-tasks occur in sequence without pause.

\begin{figure*}[t]
    \centering
    \begin{subfigure}[b]{0.49\textwidth}
        \includegraphics[width=\textwidth]{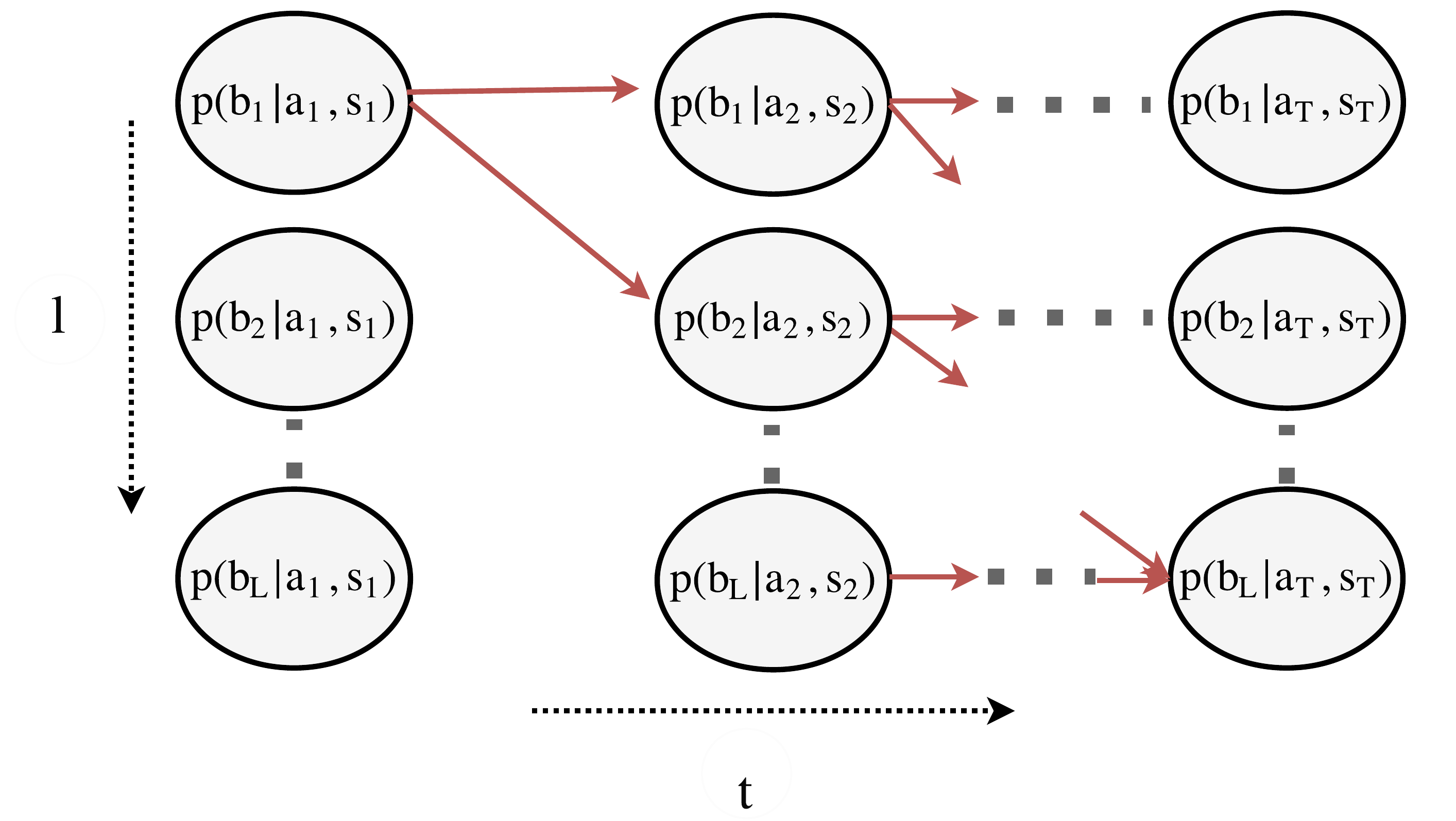}
        \caption{CTC~-~Computation of forward variables $\alpha_t(l)$}
        \label{fig:ctc_forward}
    \end{subfigure}
    ~ 
    \begin{subfigure}[b]{0.46\textwidth}
        \includegraphics[width=\textwidth]{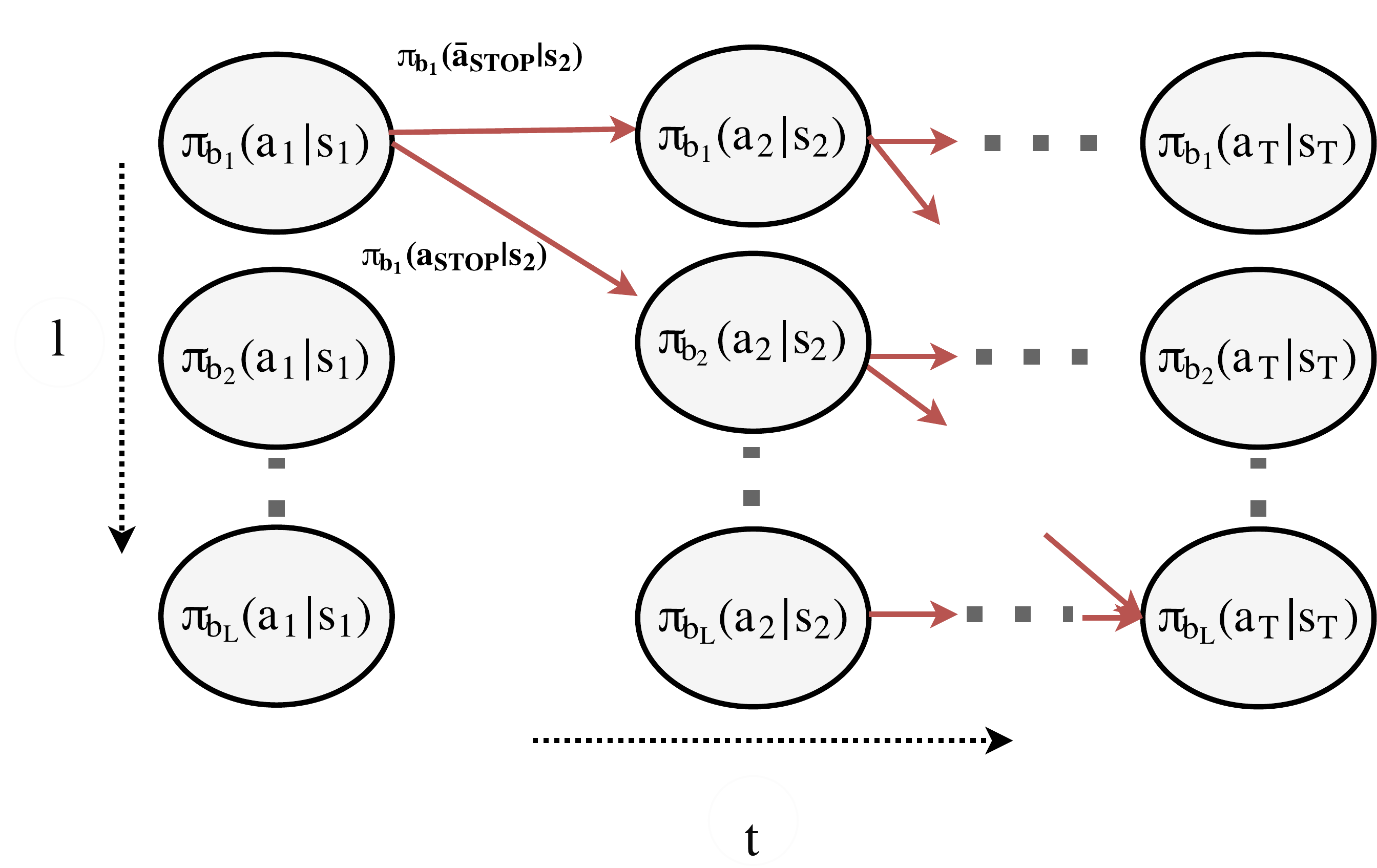}
        \caption{\method~-~Computation of forward variables $\alpha_t(l)$}
        \label{fig:taco_forward}
    \end{subfigure}
    \caption{Visualisation of the forward recursion for all paths $\zeta$ corresponding to the sketch $\tau$. The horizontal axis denotes time, while the vertical denotes the task sequence. While a node at $l,t$ in CTC is only weighted by the meta-controller probability $p(b_l|s_t,a_t)$, nodes and edges are weighted in \method~respectively via the sub-policies $\pi_l(a_t|s_t)$ and $\pi_l(a_{STOP}|s_t)$  }
    \label{fig:animals}
\end{figure*}

A \emph{path} $\zeta = (\zeta_1,\zeta_2, ..., \zeta_{T})$ is a sequence of sub-tasks of the same length as the input sequence $\rho$, describing the active sub-task $\zeta_t$ from the dictionary $\mathcal{B}$ at every time-step. 
The set of all possible paths $\mathcal{Z}_{T,\tau}$ for a task sketch $\tau$ is the set of paths of length $T$ that are equal to $\tau$ after removing all adjacent duplicates. For example, after removing adjacent duplicates, the path $\zeta=(b_1,b_1,b_2,b_3,b_3,b_3)$ equals the sketch $\tau=(b_1,b_2,b_3)$.
The CTC objective maximises the probability of the sequence $\tau$ given the input sequence $\rho$:
\vspace{-15pt}
%


\begin{align}
	 \psi^* &= \argmax_{\psi} \mathbb{E}_{(\rho,\tau)}[p_{\psi}(\tau|\rho)]\label{eq:ctc_objective}\\
	&= \argmax_{\psi}\mathbb{E}_{(\rho,\tau)}\Big[\sum_{\zeta \in \mathcal{Z}_{T,\tau}}^{} p_{\psi}(\zeta|\rho)\Big]\\
	\label{eq:ctc_objective2}
    &= \argmax_{\psi}\mathbb{E}_{(\rho,\tau)}\Big[\sum_{\zeta \in \mathcal{Z}_{T,\tau}}^{} \prod_{t=1}^{T}~p_{\psi}(\zeta_t|\rho)\Big ]
\end{align}
where $p_{\psi}(\zeta_t|\rho)$ is commonly represented by a neural network parameterised by $\psi$ that outputs the probability of each sub-task in $\mathcal{B}$.
While naively computing \eqref{eq:ctc_objective} is infeasible for longer sequences, dynamic programming provides a tractable solution. 
Let $\mathcal{Z}_{t,\tau_{1:l}}$ be the set that includes paths $\zeta_{1:t}$ of length $t$ corresponding to task sketches $\tau_{1:l}$ of length $l$, and $\alpha_t(l) = \sum_{\zeta_{1:t} \in\mathcal{Z}_{t,\tau_{1:l}}}^{}~p(\zeta|\rho)$ be the probability of being in task $b_l$ at time-step $t$ in the graph in Figure \ref{fig:ctc_forward}. The probability of a task sketch given the input sequence $p(\tau|\rho)$ is equal to $\alpha_T({L})$.

We can recursively compute $\alpha_t(l)$ based on $\alpha_{t-1}(l)$, $\alpha_{t-1}(l-1)$, and the probability of the current sub-task $p(\zeta_t|\rho_t)$.
As $\tau$ begins with a specific sub-task, the initial $\alpha$'s are deterministic and the probability of starting in the corresponding policy $b_1$ at time-step $t=1$ is 1. Figure \ref{fig:ctc_forward} depicts the recursive computation of the forward terms which is mathematically summarised as:
\begin{align}
\alpha_t(l) &= p(b_l|\rho_t)
  [\alpha_{t-1}(l-1) + \alpha_{t-1}(l))]\label{eq:ctc_alpha_recursive},\\
  \alpha_1(l) &=
   \begin{cases}
      1, & \text{if}\ l=1, \\ 
      0, & \text{otherwise}.
    \end{cases}
\end{align}
Based on the recursive computation of the CTC objective in \eqref{eq:ctc_alpha_recursive} and any automatic differentiation framework, we can optimise our model. For the manual derivation of the gradients and CTC backwards variables, please see the work of \citet{graves2006connectionist}.


\section{Methods}\label{sec:methods}
This section describes two ways to apply insights from CTC to address modular LfD. We first describe a naive adaptation which performs modular LfD with arbitrary differentiable architectures and discuss its drawbacks. We then introduce TACO, which simultaneously optimises the alignment between trajectory $\rho$ and task sequence $\tau$ and learns the underlying policies.


\subsection{Naively Adapted CTC}\label{sec:ctc_naive}
A naive modular LfD algorithm can first align the state-action trajectories with the prescribed sketches via CTC to derive the required datasets $\mathcal{D}_k$ and then learn $K$ modular policies with behavioural cloning as in \eqref{eq:mLfD2}, an approach we denote as CTC-BC. 

However, this approach fundamentally differs from the regular application of CTC. At inference time, CTC is typically given a new trajectory $\rho$ and computes the most likely sketch assignment $\tau$. In contrast, we use CTC to align $\rho$ with $\tau$ and subsequently train the sub-policies via BC while discarding the CTC based controller. This alignment for BC is derived via maximisation over $\alpha_t(l)$, as computed in \eqref{eq:ctc_alpha_recursive}, at each time-step, leading to the most likely path through the sequence. Following the determination of sketch-trajectory alignment, the sub-policies are optimised using \eqref{eq:mLfD2}.

While this approach only trains sub-policies via a single alignment based on the $\argmax$ of the forward variables $\alpha$ for every time-step in \eqref{eq:ctc_alpha_recursive}, we can account for the probabilistic assignment of active sub-policies by utilising a weighting based on the forward variables. However, as no aligned targets exist for the stop actions, they have to be derived based on the relations of the normalised $\alpha$'s between consecutive time-steps. A derivation of this $\alpha$-weighted version of CTC-BC can be found in the Appendix. In our experiments, we consider the direct, single alignment resulting from taking the $\argmax$ as well as the full optimisation of the joint probabilities via \method.


A crucial drawback of CTC-BC is the independent computation of the optimisation for alignment and for imitation. The alignment affects the policy optimisation as it is performed in a later step but not vice-versa. The introduction of  \method~in the next section addresses this shortcoming.



\subsection{Temporal Alignment for Control (\method)}\label{sec:taco}


Aligning the sequences $\rho$ and $\tau$ via CTC treats the index for active sub-policies as pure symbols and fails to exploit the fact that we then need to learn the respective sub-policies via BC. 
Consequently, what CTC determines to be a good alignment might result in a badly conditioned optimisation problem for the sub-policies, converging to a local minimum and demonstrating degraded performance once deployed. 

In this section, we propose \emph{Temporal Alignment for Control} (\method), in which the alignment is influenced by the performance of the sub-policies and addressed within a single optimisation procedure. Concretely, instead of maximising Equation \eqref{eq:ctc_objective} followed by Equation \eqref{eq:mLfD2}, we seek to maximise the joint log likelihood of the task sequence and the actions conditioned on the states:

%
%
%
\begin{equation}
 p(\tau,\mathbf{a}_{\rho} | \mathbf{s}_{\rho} ) = \sum_{\zeta \in \mathcal{Z}_{T,\tau}} p(\zeta|\mathbf{s}_{\rho})\prod_{t=1}^{T}\pi_{\theta{_{\zeta_t}}}(a_t|s_t) \label{eq:TACo},
\end{equation}
%
where $p(\zeta|\mathbf{s}_{\rho})$ 
is the product of the stop, $a_{STOP}$, and non-stop, $\bar{a}_{STOP}$, probabilities associated with any given path. 
The first term in \eqref{eq:TACo}  is similar to the corresponding term in  \eqref{eq:ctc_objective2} but now depends only on states. Every possible alignment $\zeta$ dictates which data within the sequence $\rho$ is associated with which sub-policy $\pi_{\theta}$, which is the second term in \eqref{eq:TACo} and corresponds to the BC objective. Maximising thus performs simultaneous alignment of $\tau$ and $\rho$ and learns the associated policies for each sub-task.

As with CTC, the sketch length is expected to be shorter than the trajectory length, $L \ll T$, which for longer trajectories renders the computation of all paths $\zeta$ in  $\mathcal{Z}_{T,\tau}$ intractable. However, the summation over paths can be performed via dynamic programming with a forward-backward procedure similar to that of CTC. Using consistent notation, the likelihood of a being at sub-task $l$ at time $t$ can be formulated in terms of forward variables:
\begin{equation}
    \alpha_t(l) \coloneqq \sum_{\zeta_{1:t} \in \mathcal{Z}_{t,\tau_{1:l}}} p(\zeta|\mathbf{s}_{\rho})\prod_{t'=1}^{t}\pi_{\theta_{\zeta_{t'}}}(a_{t'}|s_{t'}).
\end{equation} 
Here $\tau_{1:l}$ denotes the part of the sub-task sequence until $l$.

Even though not explicitly modelling the probability of a certain sub-task given the states and actions ($p(b|s,a)$), extending the policies with the stop action $a_{STOP}$ enables switching from one sub-task in the sketch to the next. This in turn allows computation of the probability of being in a certain sub-task $b_l$ of the sketch, at time $t$. At time $t=1$ we know that $\zeta_1 = \tau_1$, i.e., we always necessarily begin with the first sub-policy described in the sketch. 
\begin{equation}
  \alpha_1(l) = \label{eq:alpha_taco_0}
   \begin{cases}
      \pi_{\theta{b_1}}(a_1|s_1), & \text{if}\ l=1, \\
      0, & \text{otherwise}.
    \end{cases}
\end{equation}
Subsequently, a certain sub-policy can only be reached by staying in the same sub-policy or stopping the previous sub-policy in the sketch using the $a_{STOP}$ action:
\begin{eqnarray}
  \alpha_t(l) =\pi_{\theta{b_l}}(a_t|s_t)\big[\alpha_{t-1}(l-1)\pi_{\theta{b_{l-1}}}(a_{STOP}|s_t) \label{eq:alpha_taco_1}\\
  \nonumber + \alpha_{t-1}(l)(1- \pi_{\theta{b_{l}}}(a_{STOP}|s_t))\big].
\end{eqnarray}

Performing this recursion until $T$ yields the forward variables. A visualisation of the recursion is shown in Figure \ref{fig:taco_forward}. The forward variables at the end of this recursion determine the likelihood in \eqref{eq:TACo}:
\begin{equation}
  \alpha_T(L) =   p(\tau,\mathbf{a}_{\rho} | \mathbf{s}_{\rho} ).
\end{equation}
Since the computation is fully differentiable, the backward variables and subsequently the gradient of the likelihood with respect to the parameters $\theta_i$ for each policy can be computed efficiently by any auto-diff framework. 
However, as the forward recursion can lead to underflow, we employ the forward variable normalisation technique from \citet{graves2006connectionist}. 

Intuitively, the probability for the stop actions of a policy at each time-step determines the weighting of data points (state-action pairs) for all sub-policies. If a sub-policy assigns low probability a specific data-point, e.g., if at similar states it has been optimised to fit different actions, the optimisation increases the probability of the preceding and succeeding policy in the sketch for that data point, effectively influencing the probabilistic alignment. 

A potential pitfall of simultaneously optimising for alignment and control is the early collapse of the alignment objective to a single path (Figure \ref{fig:taco_forward}). This stops further exposure of the sub-policies to potential state-action pairs they would be able to fit well. To achieve sufficient exploration of different possible alignments, we use dropout \citep{srivastava2014dropout}. At every forward pass, different alignment paths are sampled, exposing the sub-policies to a wider range of data-points they could potentially fit, greatly improving performance. 

\section{Experiments}\label{sec:experiments}
We evaluate \method~ across four different domains with different continuous and discrete states and actions including image-based control of a 3D robot arm. Our experiments aim to answer the following questions.
\begin{itemize}
\setlength\itemsep{-0.20em}
	\item How does \method~ perform across a range of different tasks? Can it be successfully applied to a range of architectures and input-output representations? 
    \item How does \method~ perform with respect to zero-shot imitation  on sequences not included in the training set and task sketches of different length? 
	\item How does \method~ perform in relation to baselines including CTC-BC (Section \ref{sec:ctc_naive}) and the fully supervised approach with all trajectories segmented into sub-tasks?
	\item How does the dataset size influence the relative performance of \method~ in comparison with our baselines?
\end{itemize}

The latter questions are investigated by introducing a set of baselines based on Sections \ref{sec:mlfd2} and \ref{sec:ctc_naive}.
\begin{itemize}
\setlength\itemsep{-0.20em}
    \item GT-BC, which uses ground-truth, segmented demonstrations and performs direct maximum likelihood training to optimise the sub-policies (Equation \ref{eq:mLfD2}).
    
    \item CTC-BC, which uses both bidirectional \emph{gated recurrent units} \citep{cho2014learning} [CTC-BC(GRU)] or \emph{multi-layer perceptrons} [CTC-BC(MLP)] to perform CTC. In both cases, we apply MLPs for the sub-policies.
\end{itemize}

While the options framework presents a common approach to hierarchical LfD \citep{fox2017multi,krishnan2017ddco,2017arXiv170906683H}, it predominantly neglects the possibility of additional control information to switch between different high-level tasks, as is given by task sketches. 
Given the resulting limitation of independent modelling of different high level tasks, these framework cannot efficiently model multiple task sketches. This will result in highly degraded performance when evaluation is based on multiple different tasks, like the ones given in our evaluation (in both regular and in particular zero shot scenarios). For this reason, the evaluation focuses on approaches that utilise the task sketches. 

The main evaluation metric is the task accuracy, i.e., the ratio of full tasks completed to the total attempted. In addition, the sub-task accuracy is ported to provide more insight. Finally, Table \ref{tab:seq_alignment} provides results comparing the sequence alignment accuracy. That is, the percentage of timesteps that the demonstrated trajectory was given a correct sub-task label when compared to the ground truth alignment.

We focus on testing in a zero-shot setting: the task sequences prescribed at test time are not in the training data and are longer than the training sketches. Note that in the non-zero-shot setting, while the tasks to be completed have been seen, the world parameters such as feature positions are randomised. Finally, in all our evaluations we vary the size of the training set to investigate how performance varies with respect to available data.





\subsection{Nav-World Domain}\label{sec:navworld_domain}

\begin{figure}[ht]
    \centering
    \includegraphics[scale = 0.15]{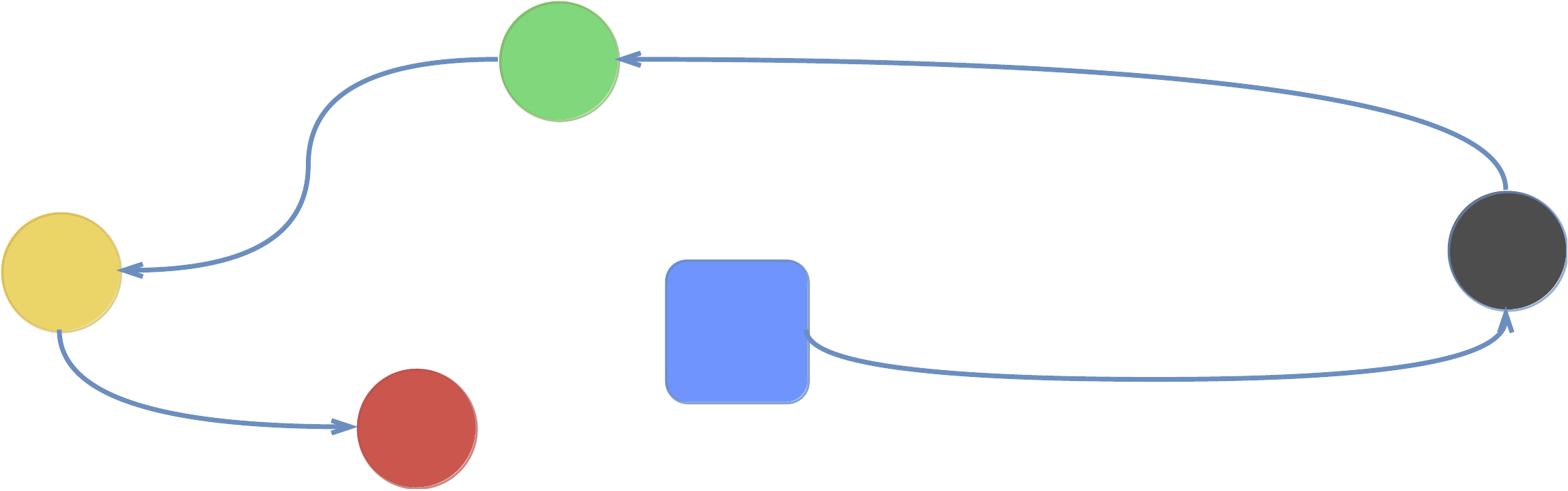}
    \caption{The Nav-World. The agent (Blue) receives a route as a task sketch. In this case. $\tau = (\text{Black, Green, Yellow, Red})$\label{fig:navworld}}
\end{figure}

We present the Nav-World domain, depicted in Figure \ref{fig:navworld}, as a simple 2D navigation task. The agent (Blue) operates in a 8-dimensional continuous state space with four destination points (Green, Red, Yellow, Black). The state space represents the $(x,y)$ distance from each of the destination points. The action space is 2-dimensional and represents a velocity $(v_x,v_y)$. At training time, the agent is presented with state-action trajectories $\rho$ from a controller that visits $L$ destinations in a certain sequence given by the task sketch, e.g. $\tau = (\text{Black, Green, Yellow, Red})$. At the end of learning, the agent outputs four sub-policies $\pi(a^+|s)$ for reaching each destination. 
At test time, it is given a sketch $\tau_{test}$ of length $L_{test}$ containing a sequence of destinations. The task is considered successful if the agent visits all destinations in the correct order. During demonstrations and testing, the agent's location and the destination points are sampled from a Gaussian distribution centred at predefined locations.
\begin{figure}[ht]
    \centering
    \includegraphics[scale = 0.25]{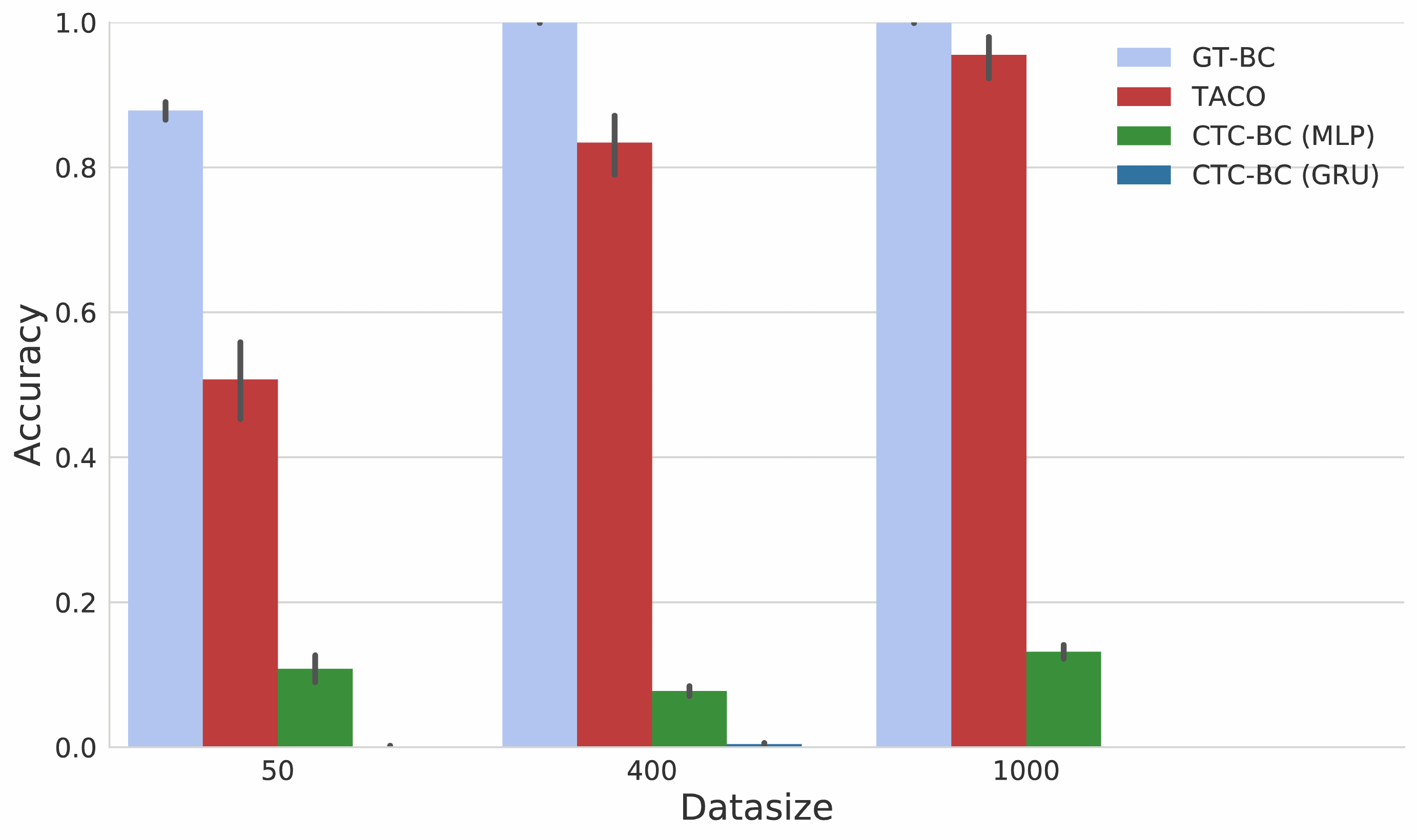}
    \caption{Nav-World results: Mean accuracy over 100 agents on 100 test tasks. Task length at test time is $L_{test}=4$ and at training time is $L=3$. TACO (red) approaches the performance of a fully supervised sequence (grey) given enough data. }
    \label{fig:navworld_results}
\end{figure}

In this domain, the dataset sizes are 50, 400, and 1000 demonstrations. The task length at training time is $L=3$. We report the task success rate for unseen, longer tasks of length $L_{test}=4$ (zero-shot setting). 100 agents are trained for each task and each algorithm, with the evaluation based on 100 testing tasks.

As displayed in the results in Figure \ref{fig:navworld_results}, 
CTC-BC performs poorly in both settings, with the MLP architecture performing slightly better. However, CTC(MLP) provides accurate alignment, often reaching 90\% overlap between the predicted sub-task sequence and ground truth (Table \ref{tab:seq_alignment}). Fitting on smaller datasets however strongly reduces the quality of the policies. As the states are similar before and after a policy switch but the actions are different (different policies), small mistakes in alignment cause multi-modality in the data distribution. This causes relatively low performance for BC with a mean squared error (MSE) objective. In contrast, \method~ avoids this problem by probabilistically weighting all policies when fitting to each time-step. From Table \ref{tab:seq_alignment} we can see that \method's~ inductive bias for control allows it to achieve much better alignment accuracy than CTC-BC. These two factors bring about performance that approaches that of GT-BC for larger datasets, a consistent trend in all our experiments.

\begin{table}[h]
    \centering
    \scriptsize    
    \begin{tabular}{l|c|c|c|c|c|c}
        & \multicolumn{4}{|c|}{Domain}\\
        \toprule
          Algorithm & Nav-World & Craft & Dial & Dial (Image)\\
        \midrule
        TACO & \textbf{95.3} & 95.6 & \textbf{99.8} & \textbf{99.0} \\
        CTC-BC (MLP) & 89.0 & 41.4 & 90.1 & 84.6 \\
        CTC-BC (GRU) & 80.0 & 57.1 & 93.4 & 48.8 \\
        GT-BC (aligned w. TACO)& 94.6 & \textbf{99.4} & 97.3 & 98.2 \\

        \bottomrule
    \end{tabular}
    \caption{Alignment accuracy of each algorithm for all domains. TACO always outperforms CTC emphasising the importance of maximising the joint likelihood of task sequences and actions. The alignment for GT-BC was obtained by computing the $\argmax$ of the TACO forward variables on the policies learned using GT-BC.}
    \label{tab:seq_alignment}
\end{table}
%
%
%
\subsection{Craft Domain}\label{sec:craft_domain}
\citet{andreas2017modular} introduced the Craft Domain to demonstrate the value of weak supervision using policy sketches in the RL setting. In this domain, an agent is given hierarchical tasks and a sketch description of that task. The binary state space has 1076 dimensions and the action space enables discrete motions as well as actions to pick up and use objects. A typical example task is $\tau_\text{make planks} =(\text{get(wood), use(workbench)})$. 
The tasks vary from $L=2$ to $L=4$. The demonstrations are provided from a trained RL agent, which obtains near optimal performance. 
Performance is measured using the reward function defined in the original domain. We train agents for all baselines and deploy them on randomly sampled tasks from the same distribution seen during training. 

\begin{figure}[ht]
    \centering
    \includegraphics[scale = 0.25]{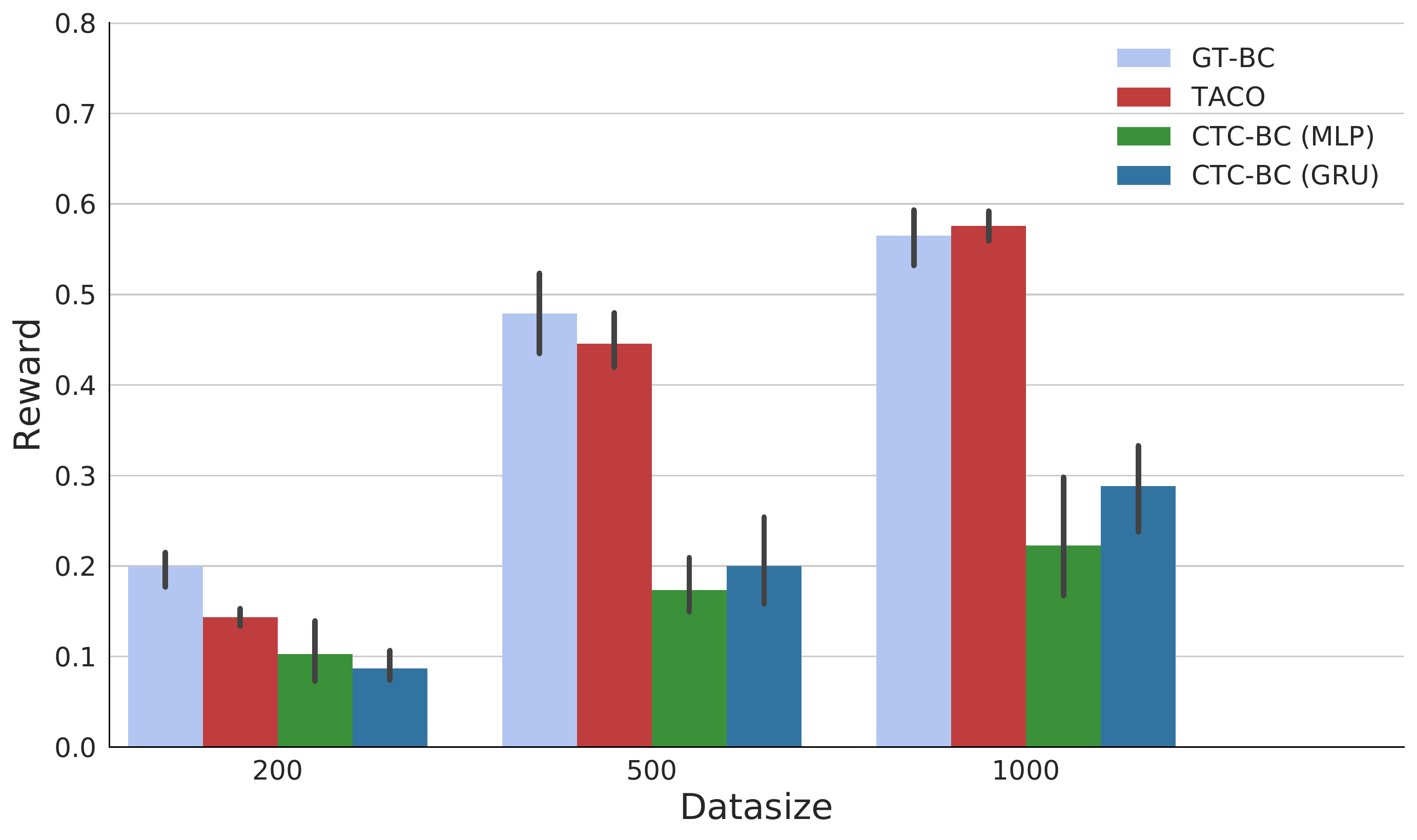}
    \caption{Craft results: Mean reward over 2000 tasks. The tasks are the same as during training but the environment instantiation is different. The RL agent used to derive the policies achieves a mean reward of 0.9. Taco (red) approaches the performance of a fully supervised method (grey) given enough data.}
    \label{fig:craftworld}
    \vspace{-0.3cm}
\end{figure}

Figure \ref{fig:craftworld} shows the results, which are similar to those in Nav-World. CTC-BC fails to obtain substantial reward. However, unlike in Nav-World, CTC does not achieve good alignments with either architecture, as the abstract, binary state-action space makes it harder to detect distribution changes. Instead of concatenating state-action pairs, \method~ learns a mapping from one to the other, making it easier to tell when the sub-policy changes. This in turn allows \method~ to match or even surpass BC with sufficient data. 

\subsection{Dial Domain}\label{sec:robot_domain}
\begin{figure}[ht]
    \centering
        \includegraphics[width=.24\textwidth,trim={0 4cm 0 0},clip]{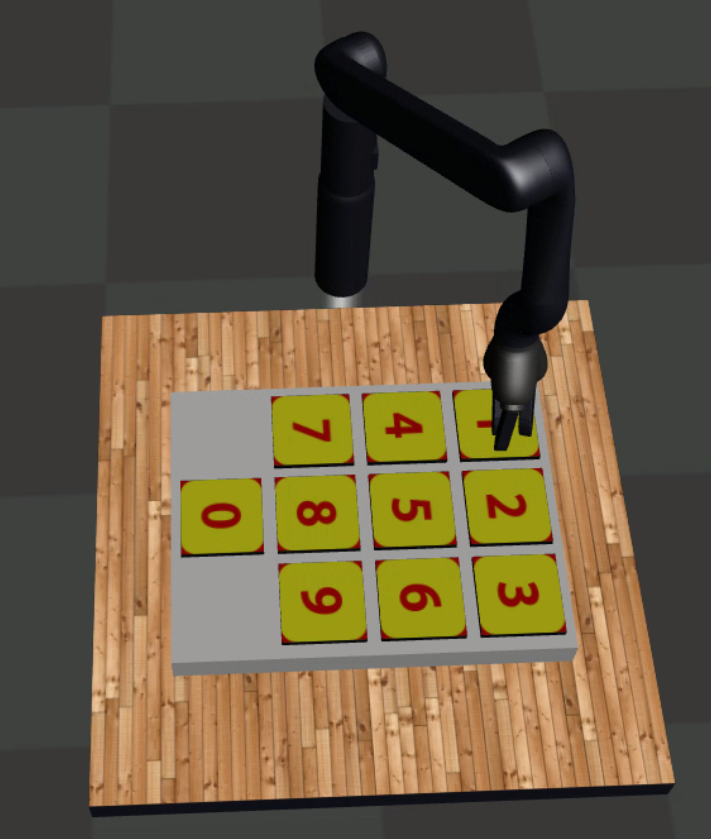}

    \caption{Dial Domain: A 6 DoF JACO arm must dial a PIN  of arbitrary length.\label{fig:dial}}
    \vspace{-0.0cm}
\end{figure}

Even though the Craft Domain is quite challenging, the lengths of the demonstrations and the resulting episodes are quite short, $T \leq 20$, making it easier to align $\rho$ and $\tau$. The final two experiments take place in a more realistic robotic manipulation domain. In the Dial Domain, a JACO 6 DoF manipulator simulated in MuJoCo \cite{todorov2012mujoco} interacts with a large dial-pad, as shown in Figure \ref{fig:dial}. The demonstrations contain state-action trajectories that describe the process by which a PIN is pressed, e.g., $\tau = (0,5,1,6)$. 
A task is considered successful if all the digits in the PIN are pressed in order. Demonstrations in this domain come from a PID controller that can move to predefined joint angles. We sub-sample the data from the simulator by 20, resulting in trajectories $T \approx 200$ for $L=4$. We report task and alignment accuracy as before. We consider two variants of the Dial Domain, one with joint-angle based states and the other with images. The action space represents the torques for each joint of the JACO arm in both cases.
\subsubsection{Joint Space Dial Domain}
In the first variant, the state is manually constructed and 39-dimensional, containing the joint angles and the distance from each digit on the dial pad in three dimensions. If the locations of the numbers in the dial-pad remains the same, however, the problem can be solved only using joint angles. For this reason, during each demonstration we randomly swap the location of the numbers. We test the policies on the standard number formation displayed in Figure \ref{fig:dial}, which is never observed during training.

\begin{figure}[ht]
    \centering
    \includegraphics[scale = 0.24]{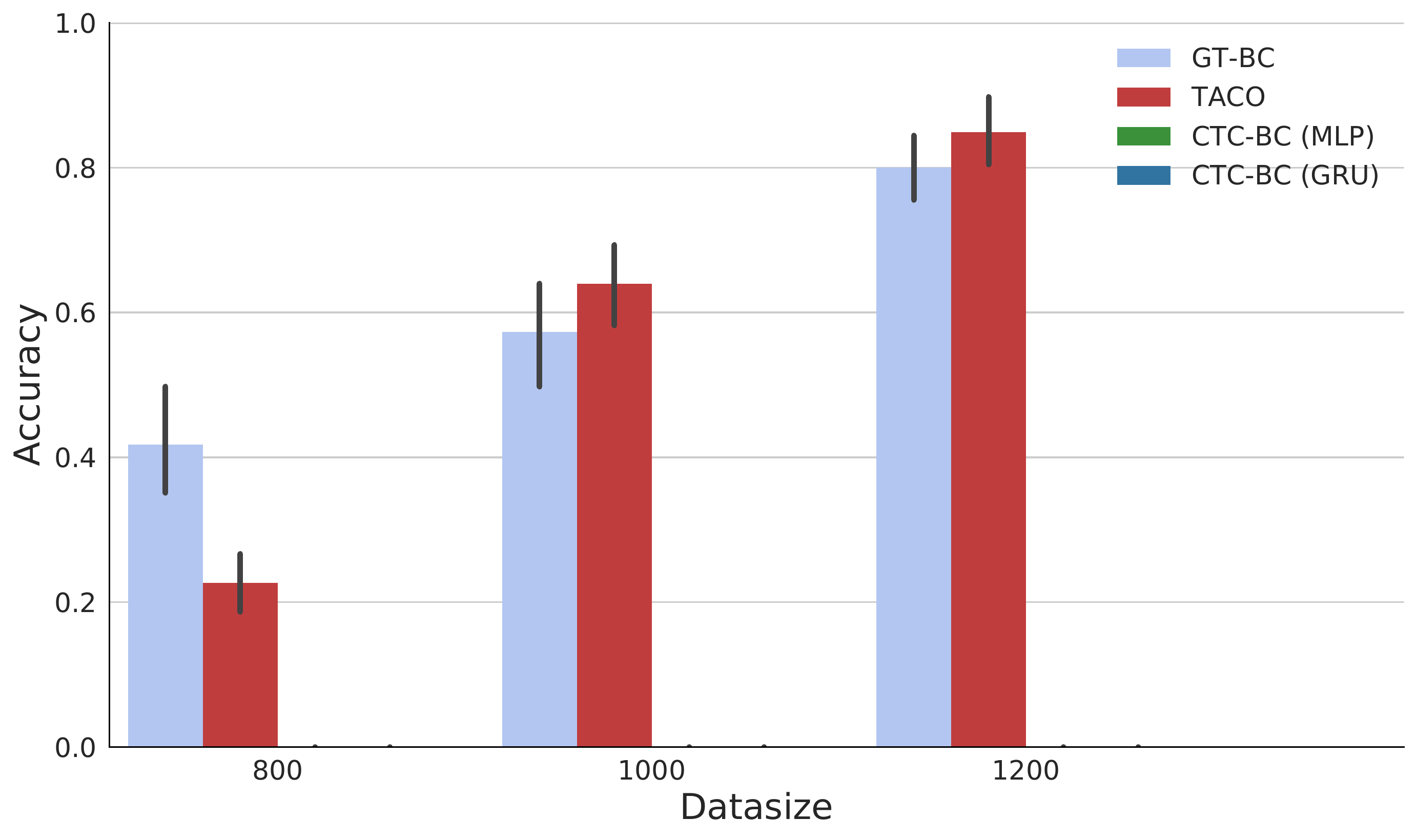}
    \vspace{-0.2cm}
    \caption{Joint Space Dial results: Task accuracy over 100 tasks of $L_{test}=5$. During training $L=4$. Evaluation is performed in an unseen configuration of the dial pad. Bars for CTC based methods do not appear as they did not finish any tasks. \label{fig:state_dial_results}}
    \vspace{-0.3cm}
\end{figure}

Figure \ref{fig:state_dial_results} shows the results, which follow the same trend as in the other domains. CTC-BC fails to complete any tasks and is not capable of aligning the sequences. TACO's performance significantly increases with data size, achieving superior performance to GT-BC at sizes of 1000 and 1200 demonstrations. We believe is due to a regularising effect of TACO's optimisation procedure. To shed more light into this observation, we use a GT-BC policy and perform alignment with the TACO forward pass on 100 unseen trajectories (Table \ref{tab:seq_alignment}). The resulting alignment is lower than that of \method, which suggests that GT-BC is more prone to overfitting. 

\begin{figure}[ht]
    \centering
    \includegraphics[scale = 0.24,trim={0cm 0.1cm 0cm 2cm},clip]{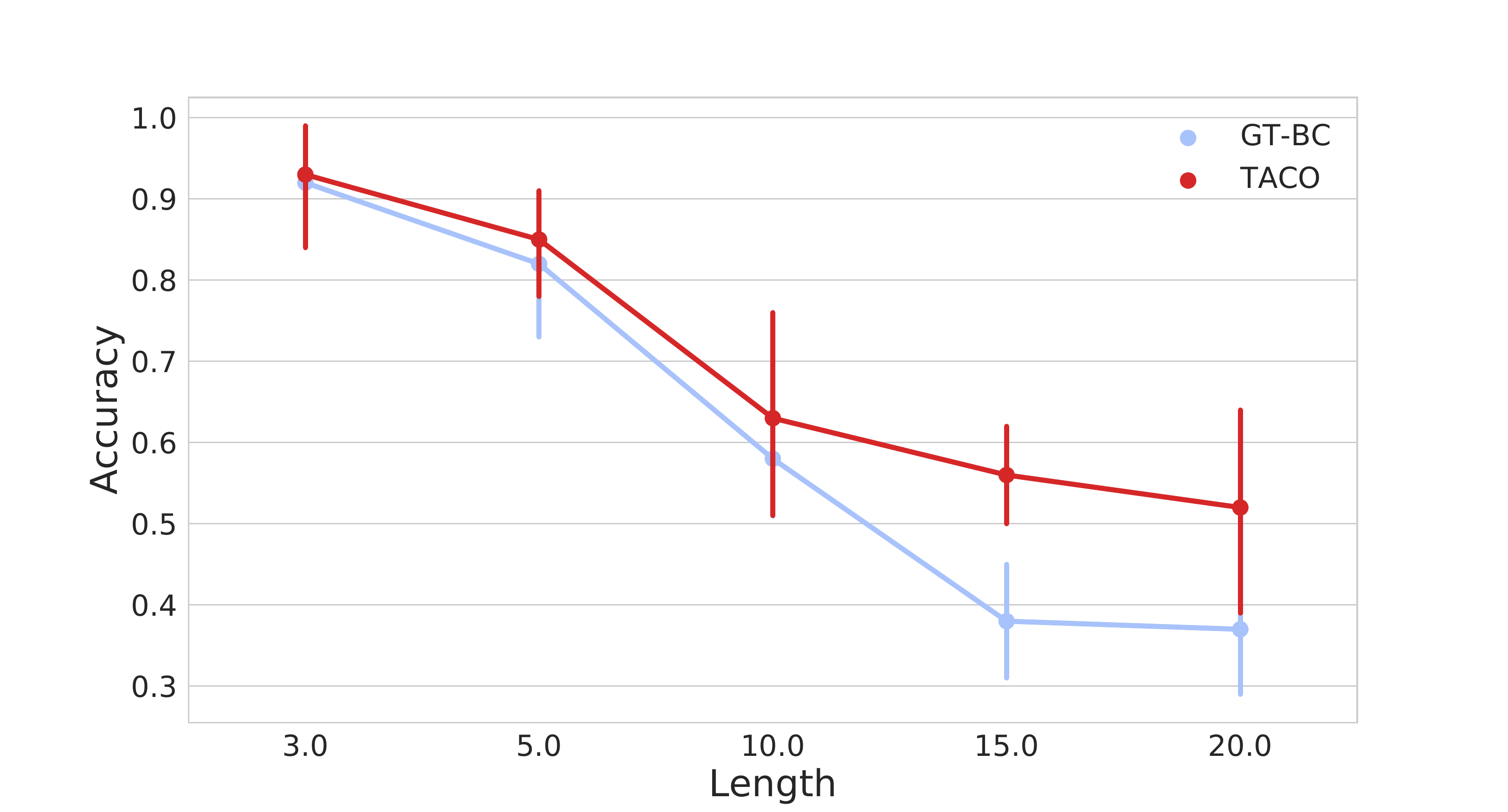}
    \vspace{-4pt}
    \caption{Task accuracy for increasing values of $L_{test}$. The accuracy is measured over 100 runs for \method~ and GT-BC.
    \label{fig:length_vs_accuracy}}
    \vspace{-0.5cm}
\end{figure}

To evaluate our methods in more complex zero-shot scenarios, we also measure the task accuracy over 100 tasks as $L_{test}$ is increased, as shown in Figure \ref{fig:length_vs_accuracy}. As expected, \method's performance falls with increasing task length as the chance of failing at a single sub-task increases. The accuracy however decreases at a lower rate than that of the baseline.

\subsubsection{Image Space Dial Domain}
The image-based variant of the Dial Domain considers the same task as above, but with an image-based state representation. We use RGB images of size $112\times112$, which are passed through a simple convolutional architecture before splitting into individual policies.  
In this case, we do not randomise the digit positions but discard joint angles from the agent's state space, as those angles would allow an agent to derive an optimal policy without utilising the image. Details of the architectures used can be found in the Appendix.

Figure \ref{fig:visual} details task and sub-task accuracy for this domain. We can see that \method~ performs well in comparison with the baselines, despite the increased difficulty in the state representation. 

\begin{figure}[h]
    \centering
    \begin{subfigure}[b]{0.21\textwidth}
    \vspace{-10pt}
        \includegraphics[height=1.1\textwidth,trim={0 0 6cm 0},clip]{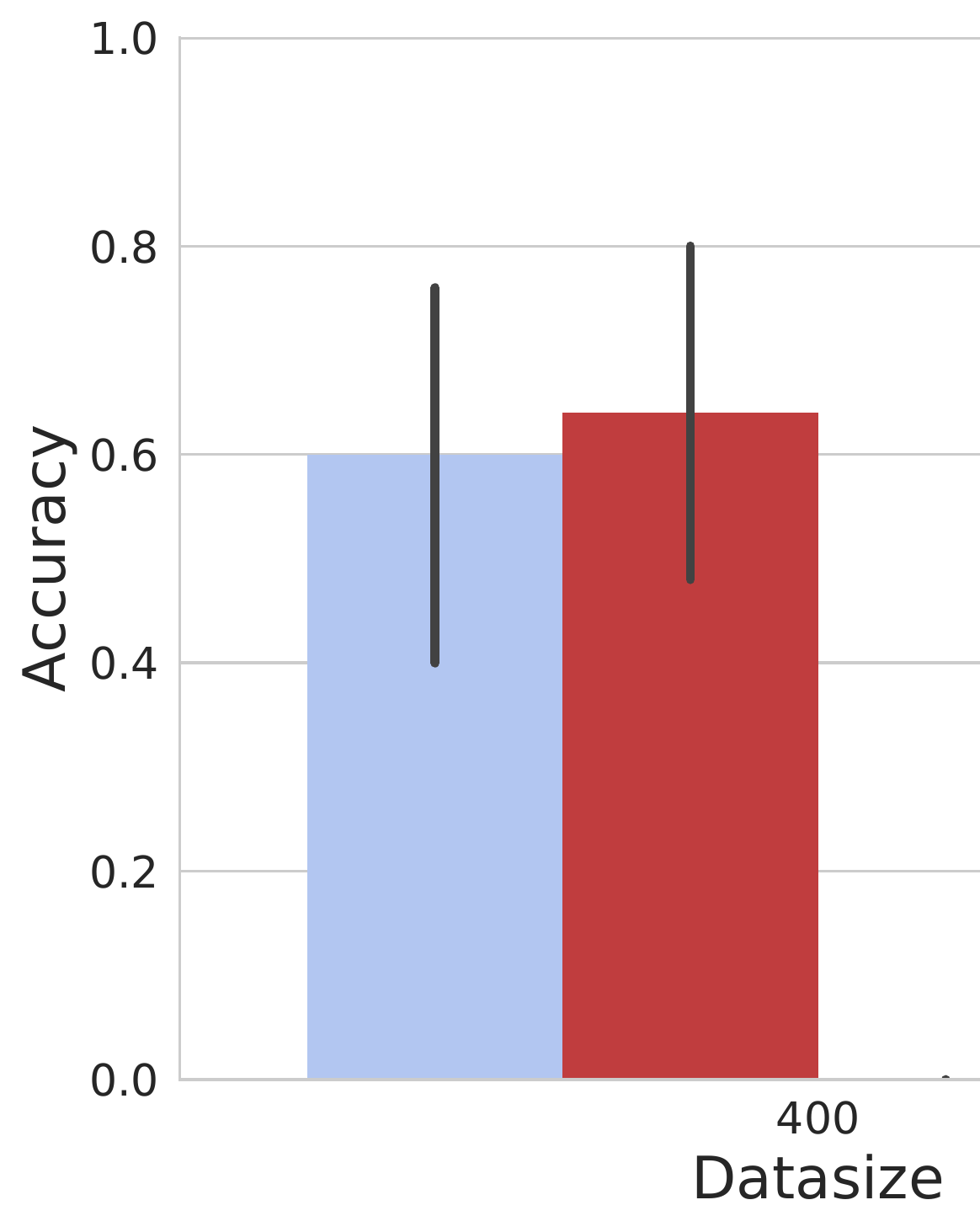}
        \caption{Task Accuracy}
        \label{fig:visual_task}
    \end{subfigure}
    \begin{subfigure}[b]{0.21\textwidth}
        \includegraphics[height=1.1\textwidth,trim={3cm 0 0 0},clip]{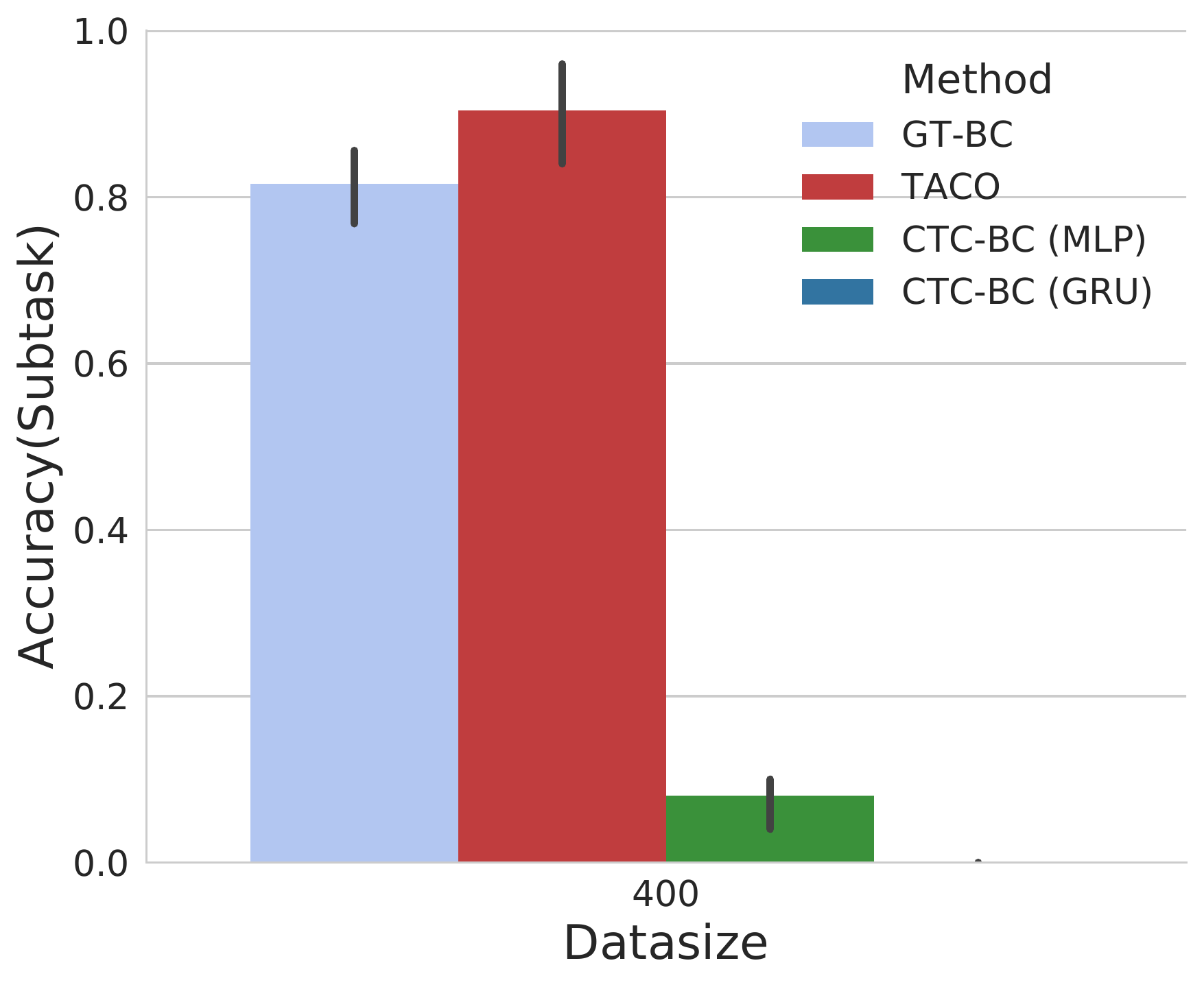}
        \caption{Sub-Task Accuracy}
        \label{fig:visual_subtask}
    \end{subfigure}
    \vspace{-10pt}
    \caption{Image-Space Dial: Accuracy over 25 test tasks. CTC-BC is unable to perform full task sequences and only solves a minor percentage of the sub-tasks, while both GT-BC and \method~ are able to complete most sub-tasks. }
    \label{fig:visual}
    \vspace{-12pt}
\end{figure}

\vspace{-5pt}





\section{Discussion}\label{sec:discussions}
\method~requires only weak supervision for modular LfD while providing performance commensurate with fully supervised approaches on a range of tasks including zero-shot imitation scenarios.

To successfully complete a task, a policy must not only complete the sub-tasks but also terminate the active policy at the right moment.
Both GT-BC and CTC-BC rely on a single alignment between sub-tasks and demonstrations, based on ground truth and an $\argmax$ alignment respectively. 
\method's strength lies in optimising the sub-policies over a distribution instead of a point estimate of the alignment. 
Training policies over a distribution of alignments exposes the sub-policies to more data points, which induces a regularising effect.  This view is supported by the alignment accuracy results of Table \ref{tab:seq_alignment} in which we evaluate trained policies for alignment on unseen sequences (based on \eqref{eq:alpha_taco_0} and \eqref{eq:alpha_taco_1}). For larger datasets 
 GT-BC achieves less accurate alignment than \method~ (Table \ref{tab:seq_alignment}), which in turn suggests that it may be more prone to overfitting. Better alignment on test demonstrations is correlated with task accuracy. This suggests that the idea of integrating the optimisation objectives in \method~for sequence alignment as well as imitation learning could have applications in cases where the end objective is good alignment rather than control policies.

\method~ has been effectively applied to the presented tasks and significantly reduces supervision efforts. However, the given sketches are highly structured and dissimilar to natural human communication. 
An interesting avenue for future work is the combination of the increased modularity of \method~ with more flexible architectures that can handle natural language \citep{mei2016listen,chaplot2017gated}. In addition, future work aims at applications in more complex hierarchical tasks and on real robots. 

While \method~ reduces the annotation effort compared to temporally segmented trajectories, it relies on weak supervision via task sketches. 
Further work into relaxing the assumptions underlying the use of these sketches can aim at omitting the constraint on the order of the sub-tasks. 



	
	
	
\vspace{-7pt}


\section{Conclusion}\label{sec:conclusions}
We presented \method, a novel method to address modular learning from demonstration by incorporating weakly supervised information in the form of a task sketch, that provides a high-level description of sub-tasks in a demonstration trajectory. 
We evaluated \method~in four different domains consisting of continuous and discrete action and state spaces, including a domain with purely visual observations.
With limited supervision, \method~performs commensurate to a fully supervised approach while significantly outperforming the straightforward adaptation of CTC for modular LfD in both control and alignment.

\section*{Acknowledgements}
The authors would like to acknowledge the support of the UK Engineering and Physical Sciences Research Council through a studentship as well as the  European Research Council FP7 program under grant agreement no. 611153 (TERESA). Furthermore, the authors would like to thank Gregory Farquhar for helping with initial support on this work. 

\bibliography{references.bib}
\bibliographystyle{icml2018}

\clearpage

\appendix

\section{Technical Details\label{ref:details}}
\subsection{Implementation Details and Architecture}

The policies for all tasks are modelled as MLPs unless images are used. In this case, convolutional layers, shared between all sub-policies, are used to extract state features. 
In continuous domains, action probabilities are modelled as $a{\sim}\mathcal{N}(\mu,\sigma=1)$ where $\mu$ is the output of the MLP. In discrete domains, action probabilities are modelled using categorical distributions. We found that having separate networks for the domain actions and the $a_{STOP}$ action, leads to better performance, especially for the Dial domain. See Table \ref{tab:Architectures} for implementation details of the architectures used in each experiment where \textit{Core} represents the convolutional architecture that learns a feature representation of the input space before feeding it into the stop and action policies.

For larger domains we found that dropout on the $a_{STOP}$ policy greatly improved results. It serves as regulariser as well as to ensure optimising over a broad range of paths as various alignment paths are sampled when units in the network are dropped. For further performance, we exponentialy decrease the dropout rate. All models are implemented in TensorFlow \cite{abadi2016tensorflow}.

\vspace{-0 mm}
\begin{table}[h]
    \centering
    \resizebox{\columnwidth}{!}{%
    \scriptsize
    \begin{tabular}{l|c|c|c|c|c}
        \toprule
          Experiment & State Dim & Core & Stop Policy & Action Policy & Output Dim\\
        \midrule
        Nav World & 8 & - & FC [100]& FC [100] & 2 \\
        Craft & 1076 & - & FC [400,100]& FC [400,100] & 7\\
        Dial & 39 & - & FC [300,200,100]& FC [300,200,100] & 9\\
        Dial (Images) & [112,112,3] & \begin{tabular}[c]{@{}l@{}}conv{[}10,5,3{]}\\ kernel{[}5,5,3{]}\\ stride{[}2,2,1{]}\end{tabular} & FC [400,300]& FC [400,300] & 9\\

        \bottomrule
    \end{tabular}
    }
    \caption{Specification  of the architectures used in each experiment. For experiment \textit{Dial (Images)} conv[], kernel[] and stride[] represent the number of channels, dimension of square kernels and 2D strides per convolutional layer respectively}
    \label{tab:Architectures}
\vspace{-5pt}
\end{table}

\subsection{Data Collection}
As mentioned in the paper, data collection took place using DART \citep{laskey2017dart} in order to compensate for the issue of covariate shift and compounding errors during policy deployment. For the Dial domain we add noise to each joint, proportional to the maximum allowed torque, which was manually tuned. During demonstrations we add a varying degree of noise to better cover the state space resulting in more robust policies that are better able to handle suboptimality in the test domain.


\section{Detailed Results \label{ref:app_res}}
This section covers more detailed results for different domains and algorithms. For the NavWorld scnario, we report results on non-0-shot settings. For the Dial domain, we additionally address the sub-task accuracy i.e how many sub-tasks were completed out of the total attempted. Finally, we detail alignment accuracy values for the methods considered in the paper. The metric describes the percentage of correctly aligned sub-policies on a set of hold-out tasks.
\subsection{NavWorld Domain}

\begin{figure}[ht]
    \centering
    \includegraphics[width=.43\textwidth]{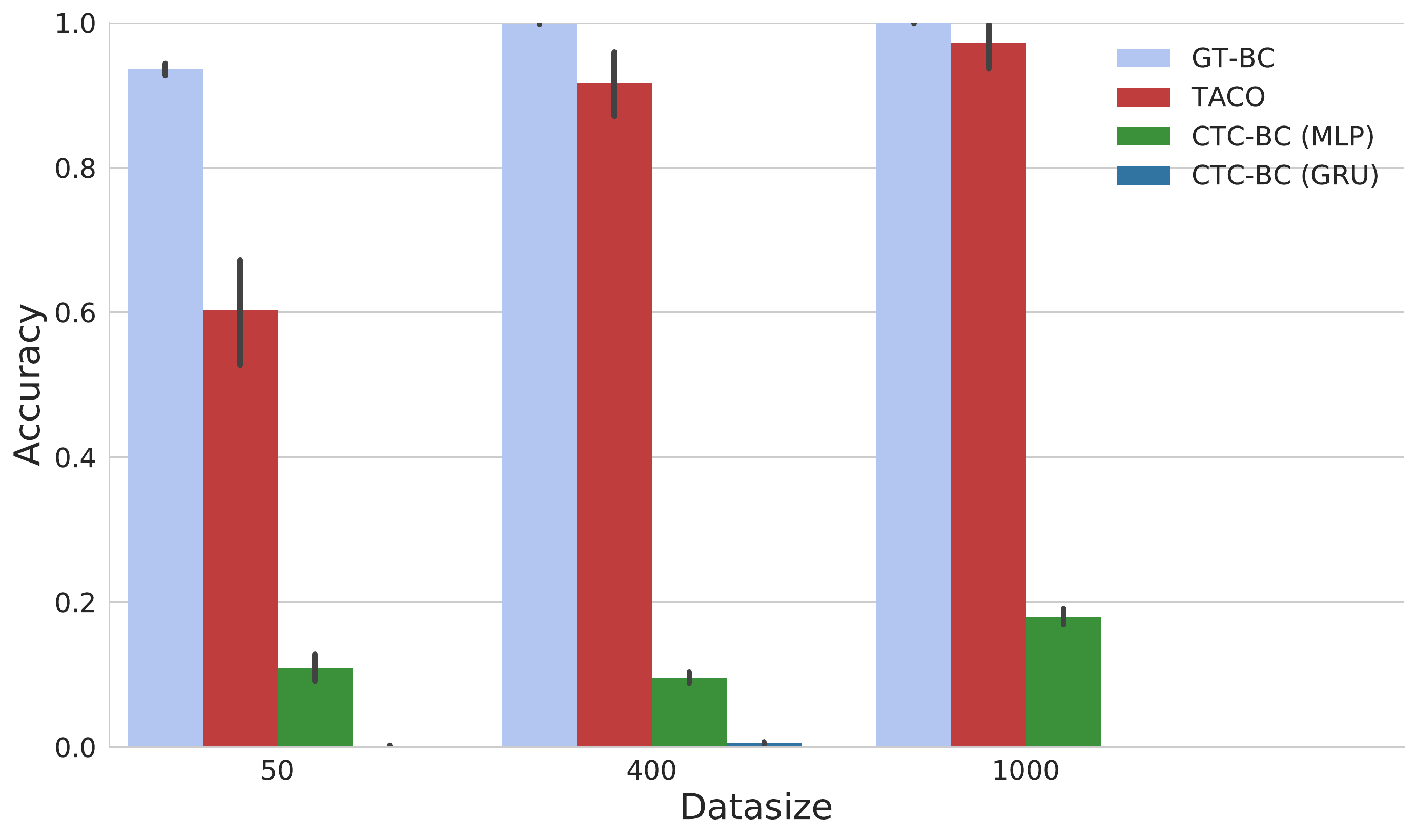}
    \caption{Task accuracy on NavWorld. $L=L_{test}=3$. }
    \label{fig:Nav_World}
\end{figure}

As is seen in Figure \ref{fig:Nav_World}, TACO not only vastly outperforms CTC-BC (MLP and GRU), both of which favour poorly in NavWorld, but as well asymptotically approaches the fully supervised GT-BC with increasing number of training trajectories.




\subsection{Joint-State Dial Domain}
\begin{figure}[ht]
    \centering
    \includegraphics[width=.43\textwidth]{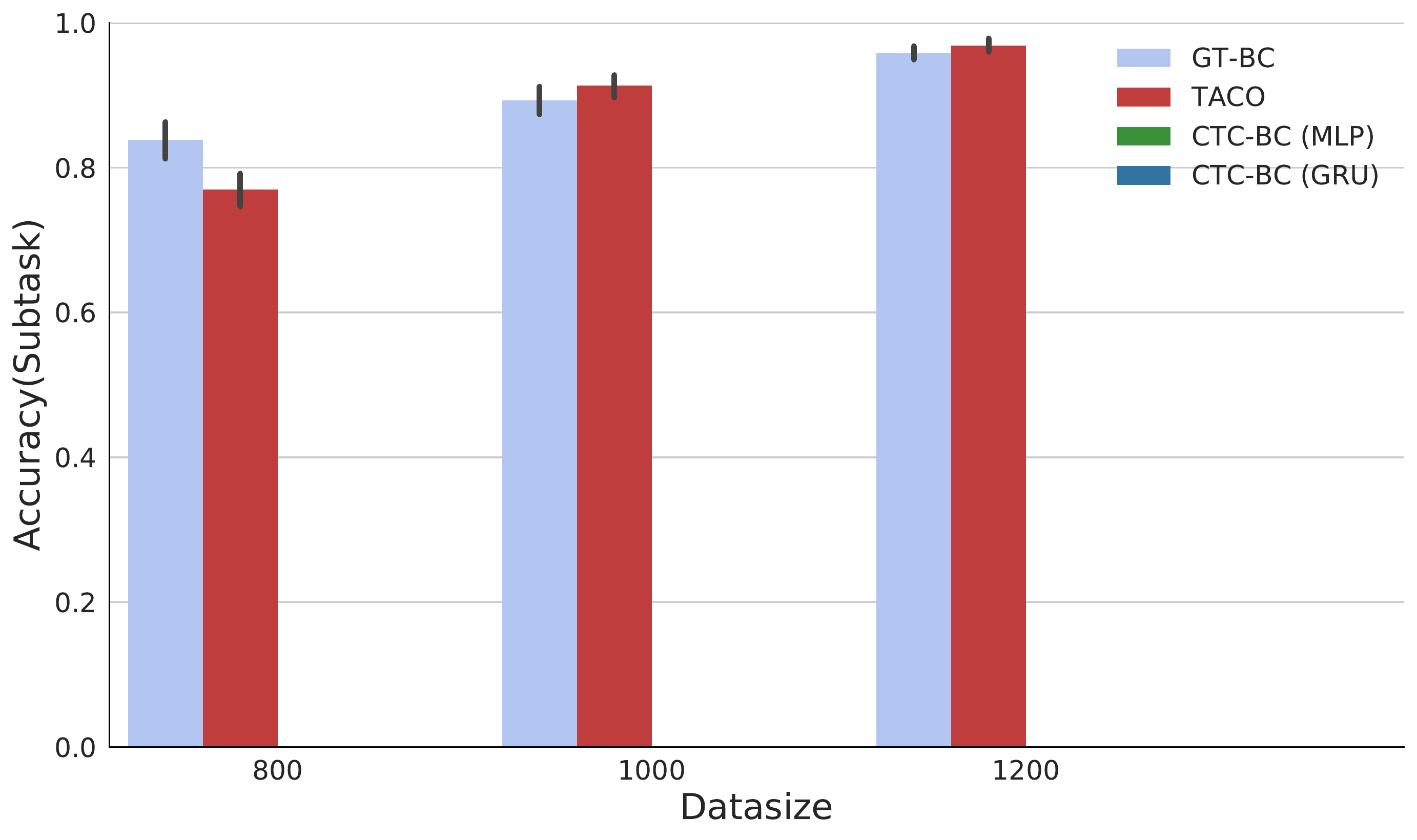}
    \caption{Sub-task accuracy for the Dial domain. We measure the number of sub-tasks succeeded over the total given, for 5 independent evaluations of 20 tasks of task length 5. Due to the complexity of the task and bad alignment performance, CTC based methods are unable to complete any sub-tasks.}
    \label{fig:j_s_d}
\end{figure}

The sub-task accuracy is displayed in Figure \ref{fig:j_s_d}.


\subsection{Sequence Alignment Accuracy}
Alignment accuracy is measured as the \% of agreement between the ground truth sub-task sequence of length $T$ and the most likely sequence predicted by either of the three alignment architectures considered in the paper, TACO, CTC-BC (MLP), CTC-BC (GRU). The most likely sequence is given by taking the $\argmax$ of the forward variables at each timestep for either CTC or TACO. For GT-BC, we pass the learned policies through TACO alignment without learning.
 We compute the alignment accuracy for 100 hold-out test sequences, which come from the same distribution as the training data. The results across all the experiment domains are stated in Table \ref{tab:seq_alignment}. 
\begin{table}[h]
    \centering
    \scriptsize    
    \begin{tabular}{l|c|c|c|c|c|c}
        & \multicolumn{4}{|c|}{Domain}\\
        \toprule
          Algorithm & Nav-World & Craft & Dial & Dial (Image)\\
        \midrule
        TACO & \textbf{95.3} & 95.6 & \textbf{98.9} & \textbf{99.0} \\
        CTC-BC (MLP) & 89.0 & 41.4 & 31.4 & 84.6 \\
        CTC-BC (GRU) & 80.0 & 57.1 & 28.6 & 48.8 \\
        GT-BC (aligned with TACO)  & 94.6 & \textbf{99.4} & 98.7 & 98.2 \\

        \bottomrule
    \end{tabular}
    \caption{Alignment accuracy of each algorithm for all domains. TACO always outperforms CTC emphasising the importance of maximising the joint likelihood of task sequences and actions.}
    \label{tab:seq_alignment}
\end{table}

\section{CTC Probabilistic Sub-Policy Training \label{ref:app_ctc}}
While the naive extension of CTC (Section 4.1) trains sub-policies based on the a single alignment by taking the $\argmax$ of the forward variables $\alpha_t(l)$ for every time-step, this paragraph addresses the possible probabilistic assignment of active sub-policies. 

To obtain the probabilistic weighting based optimisation objective in Equation \eqref{eq:mLfD_prob}, we first define the probability distribution $p_t(l)$ at time-step $t$ over all sub-policies $l$ in a sketch based on the normalised CTC forward variables in Equation \eqref{eq:ctc_pl_prob}.


However, as the ground-truth targets in the trajectory $\rho$ only exist for the regular actions, we first compute the stop action probability targets based on the CTC forward variables.

\begin{eqnarray}
	\theta_{k=1,..,K}^* = \argmax_{\theta_{k}} \mathbb{E}_{\rho_k}[\sum_{t=1}^{T}\sum_{l=1}^{L}
	\log p_t(l) \pi_{\theta_k}(a^+_t|s_t)] \label{eq:mLfD_prob}\\
	\text{where ~} p_t(l) = \frac{\alpha_t(l)}{\sum_{l_i=1}^{L}\alpha_t(l_i)}\label{eq:ctc_pl_prob}.
\end{eqnarray}
\begin{figure}[ht]
    \centering
    \includegraphics[width=.45\textwidth]{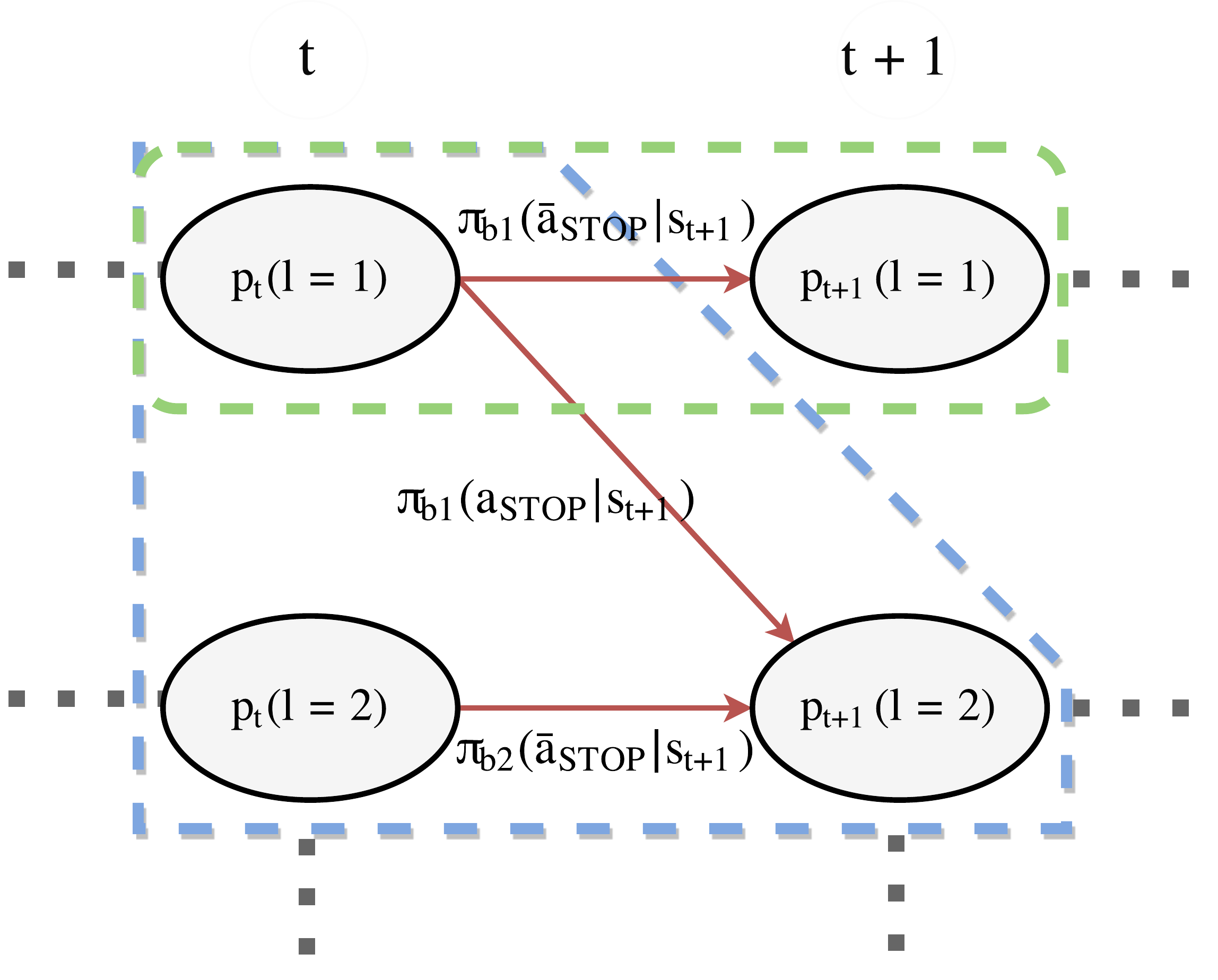}
    \caption{CTC-based computation of stop action targets. The \color{OliveGreen}{green} \color{black}{and} \color{NavyBlue}{blue} \color{black}{areas respectively depict the relations for $l=1$ in Equation \ref{eq:ctc_pl_prob_stay} and $l>1$ in Equation \ref{eq:ctc_pl_prob_stop}}}
    \label{fig:ctc_stop_props}
\end{figure}

To determine probabilistic targets for the stop actions, we associate the edges in Figure \ref{fig:ctc_stop_props} between nodes of the same or subsequent sub-policies respectively with $\bar{a}_{stop}$ and ${a_{stop}}$. For $l=1$, nodes depend only on a single relevant edge from the same sub-policy at the previous time-step, while for all nodes with $l>1$ we take into account edges from the same sub-policy and the previous sub-policy at the previous time-step.
\begin{eqnarray}
    \color{OliveGreen}{p_{t+1}(1) = p_{t}(1) \cdot \pi_1(\bar{a}_{stop}|s_{t+1})]} \label{eq:ctc_pl_prob_stay}\\
    \color{NavyBlue}{p_{t+1}(l+1) = p_{t}(l+1) \cdot \pi_{l+1}(\bar{a}_{stop}|s_{t+1}) +}\label{eq:ctc_pl_prob_stop} \\  
    \nonumber \color{NavyBlue}{p_{t}(l) \cdot \pi_{l}(a_{stop}|s_{t+1})}  \\
    \pi_l(a_{stop}|s_{t}) = 1 - \pi_l(\bar{a}_{stop}|s_{t}) \label{eq:ctc_pl_prob_negate}
\end{eqnarray}
Based on Equations \eqref{eq:ctc_pl_prob_stay}, \eqref{eq:ctc_pl_prob_stop} and \eqref{eq:ctc_pl_prob_negate}, we can derive the targets in Equation \eqref{eq:ctc_pl_prob_targets} for $t=1, ..., T-1$. Starting with the targets for $l=1$ we can compute the targets for $l+1$ based on the targets for $l$ and the forward variables $\alpha_t(l)$.  

\begin{eqnarray}
  \pi_l(\bar{a}_{stop}|s_{t+1}) \label{eq:ctc_pl_prob_targets}
  \begin{cases}
     \frac{p_{t}(l)}{p_{t+1}(l)}, & \text{if}\ l=1, \\
     \\
     \frac{p_{t+1}(l+1)}{p_{t}(l+1)} & \text{if}\ l>1.\\~~- \frac{p_{t}(l) \cdot (1-\pi_{l}(\bar{a}_{stop}|s_{t+1}))}{p_{t}(l+1)} 
  \end{cases}
\end{eqnarray}


%




\end{document}